\pdfoutput=1

\documentclass[11pt]{article}

\usepackage{EACL2023}

\usepackage{times}
\usepackage{latexsym}

\usepackage[T1]{fontenc}

\usepackage[utf8]{inputenc}

\usepackage{microtype}

\usepackage{inconsolata}

\usepackage{booktabs}
\usepackage{multirow}
\usepackage{hyperref}
\usepackage{listings}
\usepackage{multicol}
\usepackage{graphicx}
\usepackage{mdframed}
\usepackage{xcolor}
\usepackage{caption}
\usepackage{subcaption}

\definecolor{mutedgreen}{RGB}{144, 195, 141}

\title{Bryndza at ClimateActivism 2024: Stance, Target and Hate Event Detection via Retrieval-Augmented GPT-4 and LLaMA}

\author{Marek \v{S}uppa$^{\alpha\beta}$ \quad Daniel Skala$^{\alpha\gamma}$ \quad Daniela Ja\v{s}\v{s}$^{\alpha}$ \quad Samuel Su\v{c}\'{i}k$^{\alpha}$ \quad Andrej \v{S}vec$^{\alpha}$ \quad Peter Hra\v{s}ka$^{\alpha}$\\
$^{\alpha}$Cisco / Slido,
$^{\beta}$Comenius University in Bratislava,
$^{\gamma}$University of Groningen
}
\begin{document}
\maketitle
\begin{abstract}




This study details our approach for the CASE 2024 Shared Task on Climate Activism Stance and Hate Event Detection, focusing on Hate Speech Detection, Hate Speech Target Identification, and Stance Detection as classification challenges. We explored the capability of Large Language Models (LLMs), particularly GPT-4, in zero- or few-shot settings enhanced by retrieval augmentation and re-ranking for Tweet classification. Our goal was to determine if LLMs could match or surpass traditional methods in this context.

We conducted an ablation study with LLaMA for comparison, and our results indicate that our models significantly outperformed the baselines, securing second place in the Target Detection task. The code for our submission is available at \url{https://github.com/NaiveNeuron/bryndza-case-2024}.

\end{abstract}

\section{Introduction}

The Climate Activism Stance and Hate Event Detection \cite{thapa2024stance} aims to extend the growing body of work on stance, target and hate event detection \cite{hsdnlp-parihar-2021} by exploring these tasks in the context of Climate Activism. It does so by utilizing a novel ClimaConvo dataset \cite{shiwakoti2024analyzing}, which is one of the first multi-aspect datasets of its kind.

While traditional approaches to stance, target, and hate event detection rely on finetuned classifiers, our study takes a different route. We explore how a data scientist or analyst, with only API access to a Large Language Model (LLM) and without the option to finetune or alter the model, can still develop effective solutions. By creatively adjusting the prompts given to the LLM and using external tools like vector databases and pretrained ranking models for enhancement, we've found this simple method to be surprisingly competitive. Despite its simplicity, it secured the second-highest performance in the target detection subtask.

\section{Related Work}

For the past couple of years, the progress of Natural Language Processing has been driven largely by existence and availability of datasets and data resources. In the context of climate, some notable examples include Climatebert: A pretrained language model for climate-related text \cite{webersinke2021climatebert}, a dataset for detecting real-world environmental claims \cite{stammbach2022dataset} as well as the newly introduced ClimaConvo dataset \cite{shiwakoti2024analyzing}, which forms the basis of the shared task on Stance and Hate Event Detection in Tweets Related to Climate Activism.

All of the subtasks of this shared task can be modeled as classification problems and as such there exists an extensive body of academic work on this topic. In particular, methods like SVM \cite{malmasi2017detecting}, LSTM \cite{del2017hate} as well as custom architectures such as DeepHate \cite{cao2020deephate} have been proposed and evaluated. Inspired by outstanding generalizational ability of Large Language Models -- including ChatGPT \footnote{\url{https://chat.openai.com}}, GPT-4 \cite{achiam2023gpt}, LLaMA \cite{touvron2023llama} and others -- and their performance in classification tasks, especially in zero- and few-shot settings, we investigate their adaptability and effectiveness for the tasks of stance, target and hate event detection. Although works whose aim would be similar do exist, such as for instance \cite{cruickshank2023use} and \cite{guo2024investigation}, a shared task provides a unique opportunity for a thorough evaluation on many dimensions, which is lacking in the literature and uniquely distinguishes our work.

\section{Dataset}

To execute the described experiments we used the dataset introduced in \cite{shiwakoti2024analyzing} and described in \autoref{tab:dataset}. In line with the framework of shared tasks, during the ''evaluation'' stage of the shared task the organizers first shared the train split of the datasets, using the validation split for testing. When it came to the ''testing'' stage, the organizers released labels associated with the validation split, leaving the test part of the dataset for testing and final evaluation. Hence, the evaluated models had access to both the train and valid parts of the dataset.

\begin{table}[h!]
    \centering
  \resizebox{0.48\textwidth}{!}{

    \begin{tabular}{ccccc}
    \toprule
    Subtask & Classes & Train & Valid & Test \\
    \toprule
    \multirow{2}{*}{Subtask A} & \texttt{Non-Hate} & 6385 & 1371 & 1374  \\
                               & \texttt{Hate}     &  899 &  190 &  188  \\
    
    \midrule

    \multirow{3}{*}{Subtask B} & \texttt{Individual} & 563 & 120 & 121 \\
                             & \texttt{Organization} & 105 &  23 &  23 \\
                                & \texttt{Community} &  31 &   7 &   6 \\

    \midrule

    \multirow{3}{*}{Subtask C} & \texttt{Support} & 4328 & 897 & 921 \\
                                & \texttt{Oppose} &  700 & 153 & 141 \\
                               & \texttt{Neutral} & 2256 & 511 & 500  \\

    \bottomrule
    \end{tabular}
    }
    \caption{Statistics of the train, valid and test splits of the provided datset. Note that the datasets for Subtask A and Subtask B are exactly the same content-wise; it is just the labels that change.}
    \label{tab:dataset}
\end{table}

As we can observe in \autoref{tab:dataset}, the splits of the datasets are generally evenly split across the three subtasks. It seems the only exception is the Subtask C (stance detection), in which both the train and valid sets were split in 59:31:10 and 57:33:10 rations respectively, whereas the test set was split in 59:32:9 ratio.

A cursory glance at the dataset has also revealed that a relatively significant proportion of its tweets (489 in total) contains the sentence ''You've been fooled by Greta Thunberg''. While an interesting tidbit, it is almost certainly an artefact of the data collection process and provides insight into the peculiarities of the task and the data it uses for evaluation -- particularly since in an overwhelming number of cases the tweets that contain this substring are labelled as \texttt{Hate}, \texttt{Individual} and \texttt{Oppose} for Subtasks A, B and C, respectively.

\section{System description}
\label{sec:system-desc}
As outlined above, the primary component of our system is a Large Language Model, namely GPT-4, which was chosen for its strong zero-shot and few-shot capability. The model was accessed via the Azure OpenAI service and was not changed and/or finetuned as part of our experimentation -- the only attribute of the system that changed from one configuration to the other is the prompt that is sent to the GPT-4 API. In our experiments we utilized the \texttt{2023-07-01-preview} version\footnote{\url{https://learn.microsoft.com/en-us/azure/ai-services/openai/reference}} and unless otherwise noted, the temperature has been set to 0 in order to make the experiments reproducible. We also utilize paralellism in order to decrease the time necessary for the whole pipeline to run. In the end, the evaluation of our models on Subtask A and Subtask C takes roughly 25 minutes, whereas it is possible to evaluate Subtask B within 2 minutes and 30 seconds.

\subsection{Obtaining the prompt template}

As we already established, the prompt is the crucial part of our system, as it is its only changing part. To arrive at a suitable prompt for each of the subtasks, we utilized GPT-4 itself. Let us illustrate this approach on Subtask A. To generate its prompt,  a small sample of 30 \texttt{Non-Hate} and 30 Hate tweets has been selected and sent to GPT-4 along with the following prefix:

\begin{mdframed}
\begin{lstlisting}[breaklines=true,basicstyle=\ttfamily]
You will be given $n_examples tweets that were classified as hate speech. Your task is to find a common pattern these texts share and figure out why they were classified as hate speech. For a good comparison, I will also send you $n_examples non-hate speech tweets so you have something to compare it to.  Since these are tweets, focus on hashtags (#).
\end{lstlisting}
\end{mdframed}

Note further that the \texttt{\$n\_examples} in the prompt would be replaced with the actual number of examples provided after this ''prompt prefix''. The resulting response from GPT-4 would then be lightly edited by a human expert (typically done by one of the authors to ensure common formatting across all the prompts) such that the end result would be a prompt similar to that presented in \autoref{sec:prompt-subtask-a}.

\subsection{Retrieval-augmentation}

As we can see in the prompts listed in \autoref{sec:prompt-subtask-a}, \autoref{sec:prompt-subtask-b} and \autoref{sec:prompt-subtask-c}, each of the prompts (or prompt templates/prefixes) ends with a \texttt{\#\# Examples} section. This section is optional and does not necessarily need to be populated, in which case GPT-4 would be used in so called zero-shot setup (model \texttt{\textbf{GPT-4}} in \autoref{tab:experiments}). If examples are to be used, however, there are multiple options for choosing them.

The first one is to choose a fixed number of examples ($k$) that will be part of the prompt template every time it is used and will not change with each example the model processes (the \texttt{\textbf{GPT-4 few-shot}} models in \autoref{tab:experiments}). An alternative approach would be to try to extend the prompt with examples from the training set similar to the input sample in the hopes of providing further context for the LLM to make the final classification decision. This is the core idea behind retrieval-augmented generation (RAG, introduced in \cite{lewis2020retrieval}) which we adapt for our classification problems.

In particular, we utilize the Chroma vector database \footnote{\url{https://www.trychroma.com/}} to create an index of embeddings generated by one of two pre-trained Sentence Transformer models \footnote{\url{https://www.sbert.net/docs/pretrained_models.html}}: \texttt{all-MiniLM-L6-v2} which is the default embedding model the Chroma vector database makes use of and at the time of writing a Sentence Transformer with the best speed/performance ratio (resulting in the \texttt{\textbf{GPT-4 RAG}} model in \autoref{tab:experiments}) and \texttt{all-mpnet-base-v2} which reports the best peformance on standardized benchmarks at the cost of being larger and slower (and results in the \texttt{\textbf{GPT-4 RAG all}} model in \autoref{tab:experiments}). At inference time the same model that was used for index creation will provide the embedding for the sample that is being evaluated and this representation will be used to query the database, which will return the $k$ closest items from its index. These will then be lightly formatted\footnote{By ''lightly formatted'' we mean that a string denoting a beginning of the tweet would be added. There is no other pre- or post-processing done on the input data.} and provided as the final part of the prompt in the \texttt{\#\#\# Examples} section (please refer to \autoref{sec:prompt-subtask-a}, \autoref{sec:prompt-subtask-b} and \autoref{sec:prompt-subtask-c} for more details). 

Note that regardless of what process and model is being used the input tweets are used verbatim, without any pre-processing.

\subsection{Re-ranking}

Although the approach outlined in the section above is certain an improvement over a fixed list of examples, it can still potentially suffer from limitations of the underlying model(s). In particular, while they do leverage semantic information, they generally do not make use of contrastive information which in turn means that for instance the sentences ''I love trees!'' and ''I hate trees!'' will most probably have very high similarity score -- an attribute that might not be desirable in tasks like Stance, Target and Hate Event detection.

A popular way of alleviating this issue is to make use of the concept of re-ranking in which a larger number of items (for instance $3 \times k$) is requested from the index and using a pre-trained model computes relevance scores for each and thus alters their order. The top $k$ items can then be taken and processed further as described above.

In our case we use the \texttt{flashrank} library \cite{Damodaran_FlashRank_Lightest_and_2023} which provides a finetuned rank-T5-flan model based on RankT5 \cite{zhuang2023rankt5}. We also experiment with the RAGatouille library\footnote{\url{https://github.com/bclavie/RAGatouille}} but in our experiments its performance was at best comparable to that of \texttt{flashrank}, so we only report its scores in \autoref{tab:experiments} (model \texttt{\textbf{GPT-4 flashrank}}).

\subsection{Parsing the results}

As can be seen in \autoref{sec:prompt-subtask-a}, \autoref{sec:prompt-subtask-b} and \autoref{sec:prompt-subtask-c}, the prompts are designed to elicit chain-of-thought style reasoning in the model output \cite{wei2022chain}. It is hence rather difficult to ensure the output matches a specific template which would imply one of the possible classes. To that end, we match a specific keyword (e.g. \texttt{Prediction: 1}) towards both the beginning as well as the end of the LLM output.

\begin{table*}[ht!]
  \centering
  \resizebox{\textwidth}{!}{
  \begin{tabular}{lcccccccccccccccc}
    \toprule
    \multirow{2}{*}{\textbf{Model}} & \multicolumn{5}{c}{\textbf{Subtask A}} & \multicolumn{5}{c}{\textbf{Subtask B}} & \multicolumn{5}{c}{\textbf{Subtask C}}  \\
    \cline{2-13}
     & \textbf{Acc} & \textbf{P} & \textbf{R} & \textbf{F1} & \textbf{rnk} & \textbf{Acc} & \textbf{P} & \textbf{R} & \textbf{F1} & \textbf{rnk} & \textbf{Acc} & \textbf{P} & \textbf{R} & \textbf{F1} & \textbf{rnk} \\
    \midrule
    Baseline & .901 & - & - & .708 & - & .716 & - & - & .554 &  & .651 & - & - & .545 & -\\
    \midrule
    \texttt{\textbf{GPT-4}} & .935 & .835 & .880 & .856 & - & .900 & .545 & .656 & .553 & - & .693 & .515 & .513 & .509 & -\\
    
    \texttt{\textbf{GPT-4 few-shot}} (k=6) & .932 & .826 & .895 & .855 & - & .927 & \textbf{.809} & .723 & .747 & - & .693 & .502 & .507 & .487 & - \\
    \texttt{\textbf{GPT-4 few-shot}} (k=8) & .916 & .794 & .886 & .855 & - & .927 & \textbf{.809} & .723 & .747 & - & .702 & .511 & .512 & .495 & - \\
    
    \texttt{\textbf{GPT-4 RAG}} (k=4) & .944 & .859 & .890 & .874 & - & .887 & .641 & .672 & .654 & - & .707 & .517 & .514 & .498 & - \\
    \texttt{\textbf{GPT-4 RAG}} (k=6) & .941 & .851 & .889 & .868 & - & \textcolor{mutedgreen}{.927} & \textcolor{mutedgreen}{.781} & \textcolor{mutedgreen}{\textbf{.776}} & \textcolor{mutedgreen}{\textbf{.776}} & 2/18 & .690 & .668 & .681 & .666 & -\\
    \texttt{\textbf{GPT-4 RAG}} (k=8) & .942 & .855 & .887 & .870 & - & .927 & .733 & .764 & .769 & - & .688 & .666 & .678 & .661 & - \\
    
    \texttt{\textbf{GPT-4 RAG all}} (k=6) & \textcolor{mutedgreen}{\textbf{.948}} & \textcolor{mutedgreen}{\textbf{.866}} & \textcolor{mutedgreen}{\textbf{.899}} & \textcolor{mutedgreen}{\textbf{.881}} & 7/22 & .920 & .776 & .762 & .767 & - & .714 & .692 & .709 & .692 & - \\
    \texttt{\textbf{GPT-4 RAG all}} (k=8) & .944 & .864 & .884 & .874 & - & .920 & .715 & .721 & .716 & - & \textcolor{mutedgreen}{\textbf{.711}} & \textcolor{mutedgreen}{.687} & \textcolor{mutedgreen}{\textbf{.712}} & \textcolor{mutedgreen}{\textbf{.693}} & 12/19 \\
    
    \texttt{\textbf{GPT-4 flashrank}} (k=6) & .941 & .853 & .877 & .864 & - & \textbf{.940} & .635 & .617 & .625 & - & .709 & \textbf{.689} & .707 & \textbf{.693} & - \\
    \texttt{\textbf{GPT-4 flashrank}} (k=8) & .941 & .851 & .886 & .868 & - & .913 & .733 & .706 & .713 & - & .702 & .683 & .703 & .688 & - \\
    \bottomrule
  \end{tabular}
  }
  \caption{Model Performance Metrics for the respective subtasks. \textbf{Acc}, \textbf{P}, \textbf{R}, \textbf{F1} and \textbf{rnk} denote the Accuracy, Precision, Recall, the F1 score and the final rank in the Shared Task (measured by the F1 score), respectively. The final rank is reported as $r$/$n$ where $r$ denotes the position in the final results table for the respective subtask and $n$ denotes the number of teams that participated in a specific subtask. The baseline results are from \cite{shiwakoti2024analyzing}. Highest performance per each metric in each subtask is \textbf{bolded}. The performance of the model submitted to the final leaderboard is in \textcolor{mutedgreen}{green}.}
  \label{tab:experiments}
\end{table*}

\begin{figure*}[h!]
\centering
  \begin{subfigure}[b]{0.3\textwidth}
    \includegraphics[width=\linewidth]{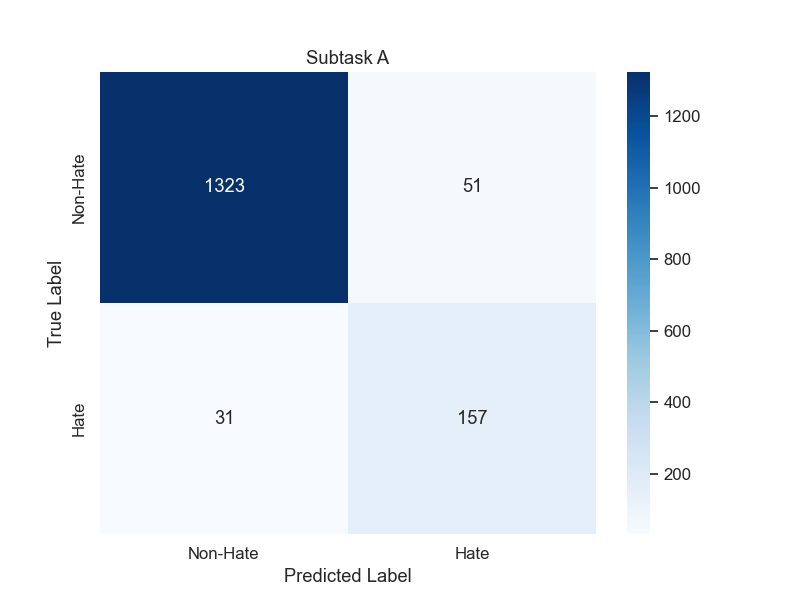}
    \caption{Confusion matrix for Subtask A}
    \label{fig:cm_subtask_a}
  \end{subfigure}
  \begin{subfigure}[b]{0.3\textwidth}
    \includegraphics[width=\linewidth]{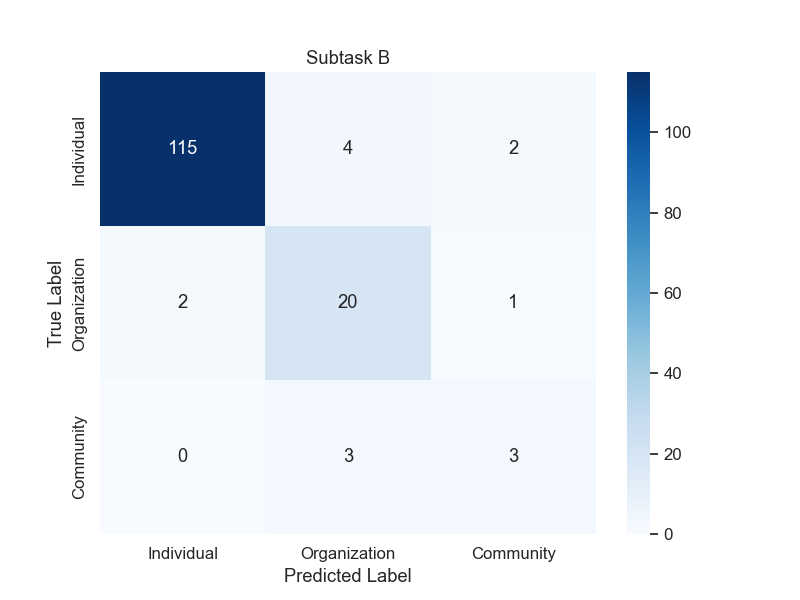}
    \caption{Confusion matrix for Subtask B}
    \label{fig:cm_subtask_b}
  \end{subfigure}
  \begin{subfigure}[b]{0.3\textwidth}
    \includegraphics[width=\linewidth]{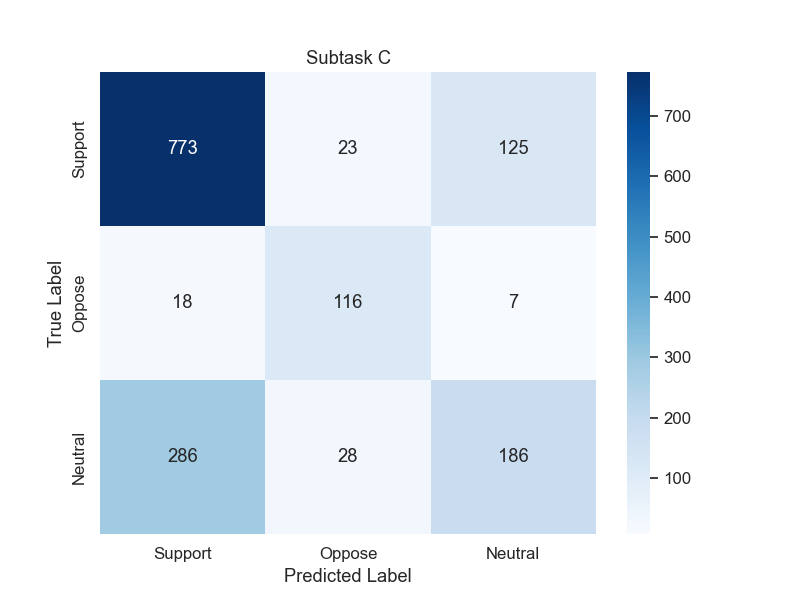}
    \caption{Confusion matrix for Subtask C}
    \label{fig:cm_subtask_c}
  \end{subfigure}
  \caption{Confusion matrices of for the best performing models on each of the subtasks.}
  \label{fig:confusion_matrices}
\end{figure*}

\section{Results \& Discussion}
\label{sec:results}

The results of our experiments can be found in \autoref{tab:experiments}. Nearly all of the models outperform the baselines introduced in \cite{shiwakoti2024analyzing} on F1 score, the primary metric chosen for this shared task. In Subtask B the baseline models report higher performance than the zero-shot evaluated GPT-4 but even a few hardcoded examples in the prompt changes the performance of the model rather dramatically (improvement of nearly 0.2 F1 points). In Subtask C we observe a similar situation, although simply adding hardcoded examples to prompt does not significantly help -- curiously enough, it even leads to decreased performance.

In general, \autoref{tab:experiments} suggest that adding retrieval agumentation generally helps, while the optimal number of examples and the optimal model in the prompt ($k$) varies per subtask. As we can see in the case of Subtask A and Subtask C, the \texttt{all-mpnet-base-v2} model has proven to be most effective, providing the final submission with $k=6$ for Subtask A and obtaining the split best performance with \texttt{\textbf{GPT-4 flashrank}} ($k=6$) in Subtask C (with $k=8$). In Subtask B the retrieval-augmentation method based on \texttt{all-MiniLM-L6-v2} yielded the best results, although the difference between the top 3 models are very small, to the point of being attributable to noise more than model/method differences. The results also suggest that the re-ranking approach using \texttt{flashrank} did not bring significant benefit over retrieval-augmentation.

In \autoref{fig:confusion_matrices} we can see the confusion matrices for the best performing models (highlighted in green in \autoref{tab:experiments}) for each of the subtasks. As the figures suggest, in Subtask A and Subtask B the models made minimal mistakes whereas in Subtask C we can observe that the model often switched the Neutral stance to Support and vice versa. We explore this phenomenon further in the next section.

\subsection{Error analysis}

To better understand the error modes of the evaluated models, we take the incorrect predictions of the best performing models and classify them into three categories: ''Error'', when the model did indeed make an incorrect prediction; ''Unclear'', when it is not clear whether the model made a mistake or whether the provided label is wrong, and ''Wrong-Label'' in which our manual annotation disagreed with that obtained from the provided test set. The annotation was done by one of the authors, followed the guidelines outlined in \cite{shiwakoti2024analyzing} and its results can be seen in \autoref{tab:error_aggregation}.

\begin{table}[h!]
\centering
\begin{subtable}{0.47\textwidth}
\centering
\caption{SubTask A: Hate Event Detection}
\resizebox{\textwidth}{!}{
\begin{tabular}{cc|cccc}
\toprule
Prediction & Label & Error & Unclear & Wrong-Label \\
\midrule
\texttt{Non-Hate} & \texttt{Hate} & 1 & 5 & 25 \\
\texttt{Hate} & \texttt{Non-Hate} & 26 & 14 & 11 \\
\bottomrule
\end{tabular}
}
\end{subtable}

\vspace{0.3em} 

\begin{subtable}{0.47\textwidth}
\centering
\caption{SubTask B: Target Detection}
\resizebox{\textwidth}{!}{
\begin{tabular}{cc|cccc}
\toprule
Prediction & Label & Error & Unclear & Wrong-Label \\
\midrule
\texttt{Individual} & \texttt{Organization} & 0 & 1 & 1 \\
\texttt{Organization} & \texttt{Individual} & 1 & 0 & 3 \\
\texttt{Organization} & \texttt{Community} & 2 & 1 & 0 \\
\texttt{Community} & \texttt{Individual} & 1 & 0 & 1 \\
\texttt{Community} & \texttt{Organization} & 0 & 0 & 1 \\
\bottomrule
\end{tabular}
}
\end{subtable}

\vspace{0.3em} 

\begin{subtable}{0.47\textwidth}
\centering
\caption{SubTask C: Stance Detection}
\resizebox{\textwidth}{!}{
\begin{tabular}{cc|cccc}
\toprule
Prediction & Label & Error & Unclear & Wrong-Label \\
\midrule
\texttt{Support} & \texttt{Oppose} & 2 & 1 & 15 \\
\texttt{Support} & \texttt{Neutral} & 10 & 8 & 268 \\
\texttt{Oppose} & \texttt{Support} & 11 & 2 & 10 \\
\texttt{Oppose} & \texttt{Neutral} & 0 & 3 & 25 \\
\texttt{Neutral} & \texttt{Support} & 46 & 12 & 67 \\
\texttt{Neutral} & \texttt{Oppose} & 0 & 2 & 5 \\
\bottomrule
\end{tabular}
}
\end{subtable}
\caption{Error type counts by Prediction and Label combinations across SubTasks. Prediction represents the model's prediction and Label the annotation obtained from the test set.}
\label{tab:error_aggregation}
\end{table}

With regards to the Hate Event Detection subtask, the model did indeed make a mistake in 27 (33\%) cases but in 36 (44\%) cases we identified a wrong label, while 19 cases (23\%) where unclear. A closer look at the error cases reveals that the model seems to overtrigger on negative concepts such as ''crimes against humanity'' or ''anger'' and considers them a \texttt{Hate} event (see \autoref{tab:errors-subtask-a}). We hypothesize that this might be an artefact of the retrieval-augmentation.

On the Target Detection subtask, the model only made 12 mistakes in total, some of which seem to stem from wrong labels (see \autoref{tab:errors-subtask-b}).

In the Stance Detection task,  a significant amount (80\%) of tweets were mislabeled, especially from \texttt{Support} to \texttt{Neutral} direction (55\%), highlighting difficulties in defining the \texttt{Support} class, like if a mention of a hashtag alone qualifies. A selection of the issues can be seen in \autoref{tab:errors-subtask-c}.

Our analysis indicates that model performance evaluation could suffer due to issues with the underlying dataset, as it contains marketing tweets irrelevant to climate activism \footnote{See the first example in \autoref{tab:errors-subtask-c}.} or single-character tweets ('0'). Had all Wrong-Label annotations been updated, the model perforamnce would be significantly higher. We recommend re-annotating the at least the test sets and updating the annotation guide to address ambiguous cases. To assist with this effort, we are releasing our error annotations as part of our submission code.

\section{Ablation study with LLaMA}

To assess to what extent would a similar approach work with a model other than GPT-4  and to provide further insight into how much of the final performance is attributable to the base model versus the other additions (e.g. RAG and/or re-ranking) we conduct an ablation study in which we replace GPT-4 with LLaMA 2 70B \cite{touvron2023llama}. We use Subtask B, in which we obtained the best results with GPT-4, as the benchmark task and due to limitations of the LLaMA's context window we further limit ourselves to $k=6$ examples in the prompt. Other than that the evaluated models are identical to those described in Section \ref{sec:system-desc}.

\begin{table}[ht!]
  \centering
  \resizebox{0.48\textwidth}{!}{
  \begin{tabular}{lccccc}
    \toprule
    \multirow{2}{*}{\textbf{Model}} & \multicolumn{5}{c}{\textbf{Subtask B}} \\
    \cline{2-6}
     & \textbf{Acc} & \textbf{P} & \textbf{R} & \textbf{F1} & \textbf{rnk} \\
    \midrule
    Baseline & .716 & - & - & .554 & - \\
    \midrule
    \texttt{\textbf{LLaMA}} & .813 & .604 & .348 & .327 & - \\
    \texttt{\textbf{LLaMA few-shot}} (k=6) & .813 & .477 & .371 & .372 & - \\
    \texttt{\textbf{LLaMA RAG}} (k=6) & .793 & .386 & .351 & .343 & - \\
    \texttt{\textbf{LLaMA RAG all}} (k=6) & \textbf{.827} & \textbf{.811} & \textbf{.482} & \textbf{.539} & 14/18 \\
    \texttt{\textbf{LLaMA flashrank}} (k=6) & \textbf{.827} & .656 & .453 & .492 & - \\
    \bottomrule
  \end{tabular}
  }
  \caption{Model Performance Metrics for the LLaMA ablation study. The legend is identical to \autoref{tab:experiments}.}
  \label{tab:llama-ablation}
\end{table}

The results can be seen in \autoref{tab:llama-ablation} where we can observe a phenomenon similar to that presented in \autoref{tab:experiments}: adding examples to the prompt generally helps, retrieval-augmentation can further improve the performance while re-ranking does not yield substantial improvement. We note, however, that comparing the two tables show that the base model has substantial impact on the final performance. In case of LLaMA, none of the evaluated models was able beat the baseline F1 score, which would land it at the 14th place (out of 18 teams). This is in direct contrast with our best model based on GPT-4, which ended up ranking second.

\section{Conclusion}

In this work we evaluate GPT-4 extended with retrieval augmentation and re-ranking on the task of Stance, Target and Hate Event Detection. We investigate to what extent is it possible utilize GPT-4 as a black box and obtain solid performance by only changing its prompt and also compare it to the LLaMA model. Our results show that augmenting GPT-4 with retrieval mechanism can lead to it serving as a classification model with competitive performance, as evidenced by the fact that it ranked 2nd in Subtask B: Target Detection.

\section*{Limitations}

The primary experiments presented in this work depend on access to a specific commercial Large Language Model which limits both the reproducibility and direct applicability of the obtained results. Similarly, the LLaMA model used in the ablation study requires a significant amount of computation resources, limiting its accessibility.

\section*{Acknowledgements}

This work was partially supported by the grant APVV-21-0114.

\bibliography{anthology,custom}
\bibliographystyle{acl_natbib}

\newpage
\appendix

\section{Prompt for Subtask A}
\label{sec:prompt-subtask-a}

\begin{lstlisting}[breaklines=true,basicstyle=\ttfamily]
Analyze the input tweet to determine if it is hate speech or not, based on the following criteria:

## Hate Speech Patterns

1. Presence of "You've been fooled by Greta Thunberg" or #FridaysForFuture in the tweet.
2. Embodies aggression or contempt towards specific groups or institutions, including dismissive attitudes towards climate activists, criticism of world leaders for climate inaction, or strong sentiments against companies investing in fossil fuels.
3. Frequent use of negative language, such as 'shame', 'lie', 'greedy', 'fake', 'idiot', to express dissatisfaction or attack others.
4. Highlights a strong ideological alignment or belief, often against fossil fuels and blaming capitalism for the climate crisis, indicating belief-driven intolerance.
5. The tone is accusatory, confrontational, and not oriented towards dialogue or understanding.

## \texttt{Non-Hate} Speech Patterns

1. Expresses concern about climate change and promotes action without aggression or contempt. Advocates for policy changes, shares environmental information, and encourages collective action rather than targeting individuals or groups.
2. Lacks negative language or personal attacks.
3. Presents a clear ideological stance on climate change in a constructive or informative manner, aiming to educate or raise awareness rather than cast blame.
4. The tone is conversational and informative, promoting understanding and engagement rather than confrontation.

## Evaluation

- If the tweet aligns more with the Hate Speech Patterns, output: 'Prediction: 1' (indicating it is hate speech).
- If the tweet aligns more with the \texttt{Non-Hate} Speech Patterns, output: 'Prediction: 0' (indicating it is not hate speech).

## Examples
\end{lstlisting}

\section{Prompt for Subtask B}
\label{sec:prompt-subtask-b}

\begin{lstlisting}[breaklines=true,basicstyle=\ttfamily]
Analyze the following tweet and classify who the target of the hate speech is. Use the identified patterns and specific examples from the training data for classification. The categories are:

## Categories

1. Individual - Involves direct attacks on specific individuals. Common examples include derogatory remarks about individuals like "Trump" or "Greta Thunberg". Look for usage of individual names and personal attacks.

2. Organization - Involves criticisms targeted at larger entities such as governments, companies, or specific organizations. Key examples include attacks on 'Government', 'Big oil companies', 'Australia' (referring to its government), 'Wilderness Committee', and the 'EU'. Look for mentions of these entities and critiques of their policies or actions.

3. Community - Involves attacks on broader communities or societal groups. Typical terms used include 'White, middle class, educated, low earners', 'humans', 'adult society', and 'politicians'. This category shifts the focus from a single party to collective human behavior, demographic groups, or societal constructs.

Use chain of thought reasoning to explain your classification. After analyzing the tweet, classify it as "Prediction: 1" for an individual, "Prediction: 2" for an organization, or "Prediction: 3" for a community. Pick only one option and put it on a new line.

## Examples
\end{lstlisting}

\section{Prompt for Subtask C}
\label{sec:prompt-subtask-c}

\begin{lstlisting}[breaklines=true,basicstyle=\ttfamily]
Analyze the following tweet and determine its stance towards the topic of Climate Activism. The stance categories are:

## Stance Categories

1. Support - These tweets show explicit support for climate action. Look for advocacy phrases like "we are mobilizing", "#ClimateJustice", "fight the #ClimateCrisis", and "Champion young people as 'drivers of change'". These often convey support through sharing news, events, or activities that promote environmental protection and sustainability.

2. Oppose - These tweets contain negative sentiments or skepticism about climate action initiatives. Phrases like "You've been fooled by Greta Thunberg", "Recycling is literally a scam!!", and rhetorical questions like "What are we saving?" are indicative of this stance. These tweets may criticize the activities of climate activists or question the credibility of climate change facts.

3. Neutral - Neutral tweets share information about climate-related activities or news without a clear stance. They use neutral language to describe events, initiatives, or outcomes, such as "At more than 750 locations worldwide - including Antarctica - youth organizers and allies united under the hashtag #PeopleNotProfit. #FridaysforFuture." These tweets do not show subjective bias or opinion towards climate action.

Keywords like 'support', 'solidarity', 'join us' suggest a supportive stance; 'fooled', 'What are we saving?', 'Greenwashing' suggest opposition; and factual reports or informative language suggest a neutral stance. The context of word usage is key for correct categorization.

Use chain of thought reasoning to explain your classification. After analyzing the tweet, classify its stance as 'Prediction: 1' for Support, 'Prediction: 2' for Oppose, or 'Prediction: 3' for Neutral. Pick only one option and put it on a new line. If the tweet is a factual statement, classify its target as described above.

## Examples
\end{lstlisting}

\section{Sample Errors}
\label{sec:errors}

\begin{table*}[ht]
    \centering
    \begin{tabular}{cc|p{10cm}}
    \toprule
    \multicolumn{3}{c}{\textbf{Wrong-Label}} \\
    \midrule
    Prediction & Label & Tweet \\
    \midrule
    \texttt{Non-Hate} & \texttt{Hate} & Young people in Bangladesh took to the streets demanding a halt to the planned expansion of the \#Matarbari coal-fired power plant. https://t.co/S5oo5Z3yCu \#FridaysForFuture \#ClimateActionNow \\
    \midrule
    \texttt{Hate} & \texttt{Non-Hate} & FFF = 666. Greta Thunberg, WEF \&amp; build back better are fronts for satan. https://t.co/uRnK9nRKIq via @YouTube \#FridaysForFuture \#GretaThunberg \#WEF \#BuildBackBetter \#Satanism \\

    \toprule
    \multicolumn{3}{c}{\textbf{Unclear}} \\
    \midrule
    \texttt{Hate} & \texttt{Non-Hate} & With every lie they’ve told, it’s our future that they’ve sold.

Week 50!! \#ClimateStrike \#FridaysForFuture \#PeopleNotProfit https://t.co/nATjq2ICKc \\
    \midrule
    \texttt{Non-Hate} & \texttt{Hate} & This \#FridaysForFuture on Zoom we will get boozy at 8pm CET (or drink soda if that's not your thing) and send some rage or wackiness to manufacturers of food items in our pantries about their packaging materials. Link information here: https://t.co/U3gdzYOcEC \#peoplenotprofit \\

    \toprule
    \multicolumn{3}{c}{\textbf{Error}} \\
    \midrule
    \texttt{Hate} & \texttt{Non-Hate} & This is huge. The top climate scientist in the world basically accuses Manchin of crimes against humanity. @s\_guilbeault @JustinTrudeau @GeorgeHeyman \#fridaysforfuture \\
    \midrule
    \texttt{Hate} & \texttt{Non-Hate} & If you are unhappy about the lack of serious climate-positive actions, put pressure on politicians. Show your anger every \#FridaysForFuture at 11 a.m. in front of Queen's Park and every other legislature and city hall in the world. Politicians are convinced that we don't care.\\
    \bottomrule
    \end{tabular}
    \caption{Sample errors annotated as part of the Error Analysis for SubTask A: Hate Event Detection.}
    \label{tab:errors-subtask-a}
\end{table*}

\begin{table*}[ht]
    \centering
    \begin{tabular}{cc|p{10cm}}
    \toprule
    \multicolumn{3}{c}{\textbf{Wrong-Label}} \\
    \midrule
    Prediction & Label & Tweet \\
    \midrule
    \texttt{Organization} & \texttt{Individual} & @Citi @Citi spent the last 5 years investing \$285 billion into destroying our futures. \#FridaysForFuture \#Divest https://t.co/y28248UskW \\
    \midrule
    \texttt{Community} & \texttt{Individual} & Wow. Blame young \#FridaysForFuture climate activists for lack of protests on the specific days of the recent heatwave, after all the vilification they've had to endure for 'skipping school'? How about some \#adultingnotadultification? \\

    \toprule
    \multicolumn{3}{c}{\textbf{Unclear}} \\
    \midrule
    \texttt{Community} & \texttt{Organization} & Week 121. Finnish forestry is bad for the climate, biodiversity and people. What Finland has is a lot of plantations and hardly any natural and old-growth forests. Finland must stop harmful forestry practices and protect and restore more forests. \#FridaysForFuture https://t.co/lLvdvlJGNh \\
    
    \toprule
    \multicolumn{3}{c}{\textbf{Error}} \\
    \midrule
    \texttt{Organization} & \texttt{Community} & @dw\_environment @Luisamneubauer @Fridays4future \#FridaysForFuture has remained influenced by strong left ideology/persons and denies the science using (existing) nuclear in climate/independence policies. \\
    \bottomrule
    \end{tabular}
    \caption{Sample errors annotated as part of the Error Analysis for SubTask B: Target Detection.}
    \label{tab:errors-subtask-b}
\end{table*}

\begin{table*}[ht]
    \centering
    \begin{tabular}{cc|p{10cm}}
    \toprule
    \multicolumn{3}{c}{\textbf{Wrong-Label}} \\
    \midrule
    Prediction & Label & Tweet \\
    \midrule
    \texttt{Neutral} & \texttt{Support} & Saasland - MultiPurpose WordPress Theme for Saas Startup: https://t.co/qbEYbFIkFy

Elementor
WooCommerce
WPML

\#WP \#WebsiteBuilder \#WebsiteDevelopment  \#100DaysOfCode \#HTML \#webdev \#WordPress \#ladningpage \#FridaysForFuture \#FridayMotivation https://t.co/4J0X5O2E3D \\
    \midrule
    \texttt{Support} & \texttt{Neutral} & Humans are destroying the very air, land and water resources we need to survive. 
\#ausvotes \#ClimateAction \#ClimateCrisis \#environment \#FridaysForFuture \#nocoal \#solarpower \#StopAdani \\

    \toprule
    \multicolumn{3}{c}{\textbf{Unclear}} \\
    \midrule
    \texttt{Support} & \texttt{Neutral} & Climate strike in Bergen, Norway. \#FridaysForFuture \#ClimateJustice \#GreenFriday @fff\_bergen https://t.co/zp4Jp6PmbP \\
    \midrule
    \texttt{Neutral} & \texttt{Support} & \#Fridaysforfuture, Dublin, Week 179. Supported by @tangfood @LoretoAbbey\_ @Janemellett @mimsmo @AngelaDeegan1 @GretaThunberg https://t.co/dtxefh9e3Y \\

    \toprule
    \multicolumn{3}{c}{\textbf{Error}} \\
    \midrule
    \texttt{Oppose} & \texttt{Support} & By no means do young people have the social \&amp; structural CAPACITIES to stand a chance against the threat that is runaway climate breakdown. Not to say that they actually did gang up and did ANYTHING in their power to deal with the problem. Look at @sunrisemvmt \&amp; \#FridaysforFuture \\
    \midrule
    \texttt{Support} & \texttt{Neutral} & Jim Cramer: Stay away from oil and gas stocks, I don't wan to touch it, stay away, no one wants oil https://t.co/Vs6DLZ1wcM , use better insulators in doors, \#fridaysforfuture, look at @Dothegreenthing https://t.co/Apxwot66Wc\\
    \bottomrule
    \end{tabular}
    \caption{Sample errors annotated as part of the Error Analysis for SubTask C: Stance Detection.}
    \label{tab:errors-subtask-c}
\end{table*}

\end{document}